\documentclass{ws-procs11x85}
\pdfoutput=1
\usepackage{ws-procs-thm} 
\usepackage{amsmath,amssymb}
\usepackage{multirow}
\usepackage[numbers, sort&compress]{natbib}
\usepackage{hyperref}       
\usepackage{url}            
\usepackage{booktabs}       
\usepackage{amsfonts}       
\usepackage{nicefrac}       
\usepackage{microtype}      
\usepackage[dvipsnames]{xcolor}
\usepackage{bbm}
\usepackage{graphicx}
\usepackage{subfigure}
\usepackage{graphicx}
\usepackage{capt-of}
\usepackage{booktabs}
\usepackage{varwidth}

\newcommand{\mtext}[1]{\text{\emph{#1}}}
\DeclareMathOperator*{\argmax}{argmax}
\usepackage[algo2e]{algorithm2e}
\usepackage{algorithm}
\begin{document}

\title{An Optimal Policy for Patient Laboratory Tests in Intensive Care Units}

\author{Li-Fang Cheng\footnotemark$^1$, Niranjani Prasad\footnotemark[\value{footnote}]$^2$ and Barbara E. Engelhardt$^{2,3}$}

\address{$^1$Department of Electrical Engineering, Princeton University \\
$^2$Department of Computer Science, Princeton University \\
$^3$Center for Statistics and Machine Learning, Princeton University}

\footnotetext{These authors contributed equally to this work.}

\begin{abstract}
Laboratory testing is an integral tool in the management of patient care in hospitals, particularly in intensive care units (ICUs). There exists an inherent trade-off in the selection and timing of lab tests between considerations of the expected utility in clinical decision-making of a given test at a specific time, and the associated cost or risk it poses to the patient. 
In this work, we introduce a framework that learns policies for ordering lab tests which optimizes for this trade-off. Our approach uses batch off-policy reinforcement learning with a composite reward function based on clinical imperatives, applied to data that include examples of clinicians ordering labs for patients. To this end, we develop and extend principles of Pareto optimality to improve the selection of actions based on multiple reward function components while respecting typical procedural considerations and prioritization of clinical goals in the ICU.
Our experiments show that we can estimate a policy that reduces the frequency of lab tests and optimizes timing to minimize information redundancy. We also find that the estimated policies typically suggest ordering lab tests well ahead of critical onsets---such as mechanical ventilation or dialysis---that depend on the lab results. We evaluate our approach by quantifying how these policies may initiate earlier onset of treatment.
\end{abstract}

\bodymatter

\section{Introduction}
Precise, targeted patient monitoring is central to improving treatment in an ICU, allowing clinicians to detect changes in patient state and to intervene promptly and only when necessary. While basic physiological parameters that can be monitored bedside (e.g., heart rate) are recorded continually, those that require invasive or expensive laboratory tests (e.g., white blood cell counts) are more intermittently sampled. These lab tests are estimated to influence up to $70\%$ percent of diagnoses or treatment decisions, and are often cited as the motivation for more costly downstream care \citep{badrick2013evidence, zhi2013landscape}.

Recent medical reviews raise several concerns about the over-ordering of lab tests in the ICU \citep{loftsgard2016clinicians}. Redundant testing can occur when labs are ordered by multiple clinicians treating the same patient or when recurring orders are placed without reassessment of clinical necessity. Many of these orders occur at time intervals that are unlikely to include a clinically relevant change or when large panel testing is repeated to detect a change in a small subset of analyses \citep{konger2016reduction}. This leads to inflation in costs of care and in the likelihood of false positives in diagnostics, and also causes unnecessary discomfort to the patient. Moreover, excessive phlebotomies (blood tests) can contribute to risk of hospital-acquired anaemia; around $95\%$ of patients in the ICU have below normal haemoglobin levels by day 3 of admission and are in need of blood transfusions. It has been shown that phlebotomy accounts for almost half the variation in the amount of blood transfused \citep{icumedical2015}.

With the disproportionate rise in lab costs relative to medical activity in recent years, there is a pressing need for a sustainable approach to test ordering. A variety of approaches have been considered to this end, including restrictions on the minimum time interval between tests or the total number of tests ordered per week. More data-driven approaches include an information theoretic framework to analyze the amount of novel information in each ICU lab test by computing conditional entropy and quantifying the decrease in novel information of a test over the first three days of an admission~\cite{lee2015using}. 

In a similar vein, a binary classifier was trained using fuzzy modeling to determine whether or not a given lab test contributes to information gain in the clinical management of patients with gastrointestinal bleeding~\cite{cismondi2013reducing}. An ``informative" lab test is one in which there is significant change in the value of the tested parameter, or where values were beyond certain clinically defined thresholds; the results suggest a $50\%$ reduction in lab tests compared with observed behaviour. More recent work looked at predicting the results of ferratin testing for iron deficiency from information in other labs performed concurrently~\cite{luosol2016}. The predictability of the measurement is inversely proportional to the novel information in the test. 
These past approaches underscore the high levels of redundancy that arise from current practice. However, there are many key clinical factors that have not been previously accounted for, such as the low-cost predictive information available from vital signs, causal connection of clinical interventions with test results, and the relative costs associated with ordering tests. 

In this work, we introduce a reinforcement learning (RL) based method to tackle the problem of developing a policy to perform actionable lab testing in ICU patients. Our approach is two-fold: first, we build an interpretable model to forecast future patient states based on past observations, including uncertainty quantification. We adapt multi-output Gaussian processes (MOGPs; \cite{ghassemi2015multivariate,Cheng2017arXiv}) to learn the patient state transition dynamics from a patient cohort including sparse and irregularly sampled medical time series data, and to predict future states of a given patient trajectory. Second, we model patient trajectories as a Markov decision process (MDP). This framework has been applied to the recommendation of treatment strategies for critical care patients in a variety of different settings, from recommending drug dosages to efficiently weaning patients from mechanical ventilation \cite{nemati2016optimal, raghu2017continuous, prasad2017reinforcement}. We design the state and reward functions of the MDP to incorporate relevant clinical information, such as the expected information gain, administered interventions, and costs of actions (here, ordering a lab test). A major challenge is designing a reward function that can trade off multiple, often opposing, objectives. There has been initial work on extending the MDP framework to composite reward functions. For example, fitted Q-iteration (FQI) has been used to learn policies for multi-objective MDPs with vector-valued rewards, for the sequence of interventions in two-stage clinical antipsychotic trials~\cite{lizotte2016multi}. A variation of Pareto domination was then used to generate a partial ordering of policies and extract all policies that are optimal for some scalarization function, leaving the choice of parameters of the scalarization function to decision makers.

Here, we look to translate these principles to the problem of lab test ordering. Specifically, we focus on blood tests relevant in the diagnosis of sepsis or acute renal failure, two common conditions associated with high mortality risk in the ICU: white blood cell count (WBC), blood lactate level, serum creatinine, and blood urea nitrogen (BUN). We present our methods within a flexible framework that can in principle be adapted to a patient cohort with different diagnoses or treatment objectives, influenced by a distinct set of lab results. Our proposed framework integrates prior work on off-policy RL and Pareto learning with practical clinical constraints to yield policies that are close to intuition demonstrated in historical data.
We apply our framework to a publicly available database of ICU admissions, evaluating the estimated policy against the policy followed by clinicians using both importance sampling based estimators for off-policy policy evaluation and by comparing against multiple clinically inspired objectives, including onset of clinical treatment that was motivated by the lab results.

\section{Methods}

\subsection{Cohort selection and preprocessing}

We extract our cohort of interest from the MIMIC III database \citep{johnson2016mimic}, which includes de-identified critical care data from over 58,000 hospital admissions.
From this database, we first select adult patients with at least one recorded measure for each of 20 vital signs and lab tests commonly ordered and reviewed by clinicians (for instance, the results reported in a complete blood count or basic metabolic panel).
We further filter patients by their length-of-stay, keeping only those that were in the ICU for more than a day but less than twenty days, to obtain a final set of 6,060 patients (Table \ref{table:covariate_stats}).
\begin{table}[h]
\vspace{-2mm}
\tbl{\textbf{Data statistics of the selected cohort.} Total number of recordings, mean value and standard deviation (SD) for each covariate in the selected cohort.}
{\begin{tabular}{p{7cm} p{1.8cm} p{1.7cm} p{1.7cm}}
\hline
Covariate & Count & Mean & SD \\
\hline\hline
Respiratory Rate (RR) & 1,046,364 & 20.1 & 5.7\\
Heart Rate (HR) & 964,804 & 87.5 & 18.2\\
Mean Blood Pressure (Mean BP) & 969,062 & 77.9 & 15.3\\
Temperature, $^{\circ}$F & 209,499 & 98.5 & 1.4\\
Creatinine & 67,565 & 1.5 & 1.2\\
Blood Urea Nitrogen (BUN) & 66,746 & 31.0 & 21.1\\
White Blood Cell Count (WBC) & 59,777 & 11.6 & 6.2\\
Lactate & 39,667 & 2.4 & 1.8\\
\hline
\end{tabular}}
\vspace{-2mm}
\label{table:covariate_stats}
\end{table}

\begin{figure}[t]
    \centering
    \includegraphics[width=1.0\textwidth]{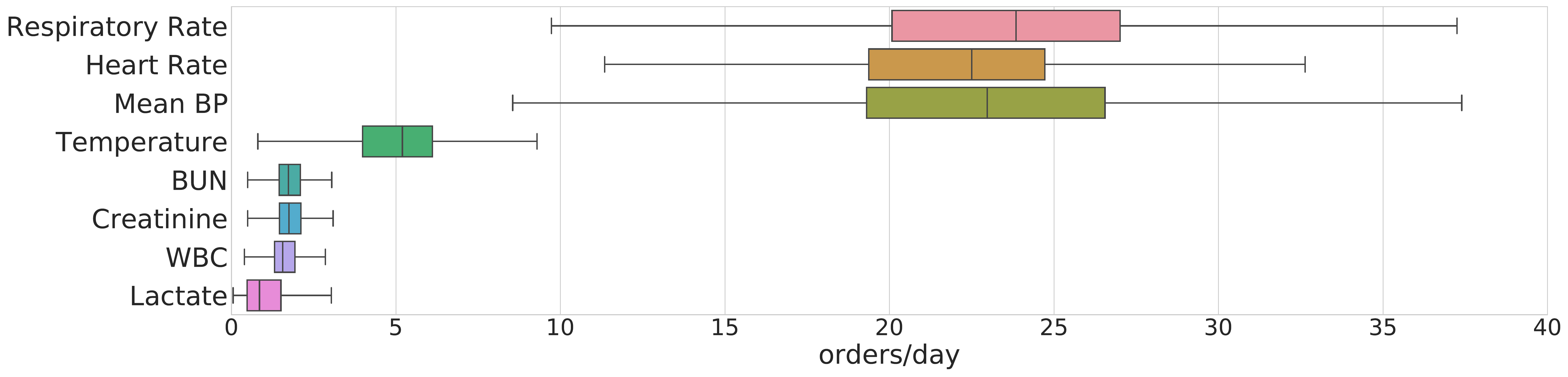}
    \vspace{-5mm}
    \caption{\textbf{Distribution of mean daily orders of the selected vitals and labs.} These eight traits are commonly used in computing clinical risk scores or diagnosing sepsis.
    }
    \label{fig:average_daily_orders}
\end{figure}

Included in the 20 physiological traits we filter for are eight that are particularly predictive of the onset of severe sepsis, septic shock, or acute kidney failure. These traits are included in the SIRS (System Inflammatory Response Syndrome) and SOFA (Sequential Organ Failure Assessment) criteria. The average daily measurements or lab test orders across the chosen cohort for these eight traits is highly variable (Figure \ref{fig:average_daily_orders}). Of these eight traits, the first three are vitals measured using bedside monitoring systems for which approximately hourly measurements are recorded; the latter four are labs requiring phlebotomy and are typically measured just 2--3 times each day. We find the frequency of orders also varies across different labs, possibly due in part to differences in cost; for example, WBC (which is relatively inexpensive to test) is on average sampled slightly more often than lactate. 
In order to apply our proposed RL algorithm to this sparse, irregularly sampled dataset, we adapt the multi-output Gaussian process (MOGP) framework \cite{Cheng2017arXiv} to obtain hourly predictions of patient state with uncertainty quantified, on 17 of the 20 clinical traits. For three of the vitals, namely the components of the Glasgow Coma Scale, we impute with the last recorded measurement.

\subsection{MDP formulation}

Each patient admission is modelled as an MDP with:
\begin{enumerate}
    \item a state space $\mathcal{S}$, such that the patient physiological state at time $t$ is given by $s_t \in \mathcal{S}$;
    \item an action space $\mathcal{A}$ from which the clinician's action $a_t$ is chosen;
    \item an unknown transition function $\mathcal{P}_{sa}$ that determines the patient dynamics; and
    \item a reward function $r_t$ that constitutes the observed clinical feedback for this action.
\end{enumerate}
The objective of the RL agent is to learn an optimal policy $\pi^*: \mathcal{S} \rightarrow \mathcal{A}$ that maximizes the expected discounted accumulated reward over the course of an admission:
\begin{equation*}
\pi^* = \argmax\limits_\pi \mathbb{E}\left[\sum\limits_{t=0}^T \gamma^t r_t | \pi\right] \text{, where $T$ is admission length, $\gamma$ is the discount factor.}
\end{equation*}
We start by describing the state space of our MDP for ordering lab tests. We first resample the raw time series using a multi-objective Gaussian process with a sampling period of one hour. The patient state at time $t$ is defined by:
\begin{equation}
\bm{s_{t}} = 
\begin{bmatrix}
{m}^{\mtext{SOFA }}_{t} &
\bm{m}^{\mtext{vitals }}_{t} &
\bm{m}^{\mtext{labs }}_{t} &
\bm{y}^{\mtext{labs }}_{t} &
\bm{\Delta}^{\mtext{labs}}_{t}
\end{bmatrix}^{\top}
\end{equation}
Here, $\bm{m}_{t}$ denotes the predictive means and standard deviations respectively of each of the vitals and lab tests. For the predictive SOFA score ${m}^{\mtext{SOFA}}_{t}$, we compute the value using its clinical definition, from the predictive means on five traits---mean BP, bilirubin, platelet, creatinine, $\mtext{FiO}_{2}$---along with GCS and related medication history (e.g., dopamine). Vitals include any time-varying physiological traits that we consider when determining whether to order a lab test. Here, we look at four key physiological traits---heart rate, respiratory rate, temperature, and mean blood pressure---and four lab tests---creatinine, BUN, WBC, and lactate. The values $\bm{y}_{t}$ are the last known measurements of each of the four labs, and $\bm{\Delta}_{t}$ denotes the elapsed time since each was last ordered. This formulation results in a 21-dimensional state space.
Depending on the labs that we wish to learn recommendations for testing, the action space $\mathcal{A}$ is a set of binary vectors whose $0/1$ elements indicate whether or not to place an order for a specific lab. These actions can be written as
$\bm{a_{t}} \in \mathcal{A} = \left\lbrace 1, 0 \right\rbrace^{L}$, where $L$ is the number of labs.

In order for our RL agent to learn a meaningful policy, we need to design a reward function that provides positive feedback for the ordering of tests where necessary, while penalizing the over- or under-ordering of any given lab test. In particular, the agent should be encouraged to order labs when the physiological state of the patient is abnormal with high probability, based on estimates from the MOGP, or when a lab is predicted to be informative (in that the forecasted value is significantly different from the last known measurement) due to a sudden change in disease state. In addition, the agent should incur some penalty whenever a lab test is taken, decaying with elapsed time since the last measurement, to reflect the effective cost (both economic and in terms of discomfort to the patient) of the test. We formulate these ideas into a vector-valued reward function $\bm{r_t} \in \mathbb{R}^d $ of the state and action at time $t$, as follows:
\begin{equation}
\bm{r_{t}} = 
\begin{bmatrix}
{r_{t}}^{\mtext{SOFA }}  &
{r_{t}}^{\mtext{treat }}  &
{r_{t}}^{\mtext{info }}  &
{-r_{t}}^{\mtext{cost}}  &
\end{bmatrix}^{\top}
\end{equation}
\paragraph{Patient state:}
The first element, $r^{\mtext{SOFA}}$, uses the recently introduced SOFA score for sepsis \cite{singer2016third} which assesses severity of organ dysfunction in a potentially septic patient. Our use of SOFA is motivated by the fact that, in practice, sepsis is more often recognized from the associated organ failure than from direct detection of the infection itself \cite{vincent2016qsofa}. The raw SOFA score ranges from 0 to 24, with a maximum of four points assigned each to symptom of failure in the respiratory system, nervous system, liver, kidneys, and blood coagulation. A change in SOFA score $\geq 2$ is considered a critical index for sepsis \cite{singer2016third}. We use this rule of thumb to design the first reward term as follows:

\begin{equation}
{r_{t}}^{\mtext{SOFA}} =\mathbbm{1}_{\bm{a_{t}} \neq \bm{0}}  \cdot\mathbbm{1}_{f(\cdot)\geq 2}\,, \text{ where } f(\cdot) = m_{t}^{\mtext{SOFA}} - m_{t-1}^{\mtext{SOFA}}.
\end{equation}
The raw score $m_{t}^{\mtext{SOFA}}$ at each time step $t$ is evaluated using current patient labs and vitals \cite{vincent2016qsofa}.

\paragraph{Treatment onset:} 
The second term is an indicator variable for rewards capturing whether or not there is some treatment or intervention initiated at the next time step, $\bm{s_{t+1}}$:
\begin{equation}
{r_{t}}^{\mtext{treat}} = \mathbbm{1}_{\bm{a_{t}} \neq \bm{0}} \cdot \sum_{i \in M} 
\mathbbm{1}_{\bm{s_{t+1}}(\text{treatment $i$ was given})},
\end{equation}
where $M$ denotes the set of disease-specific categories of interventions of interest. Again, the reward term is positive if a lab is ordered; this is based on the rationale that, if a lab test is ordered and immediately followed by an intervention, the test is likely to have provided actionable information. Possible interventions in the following state include administration of some form of antibiotics, vasopressors, initiation of dialysis or mechanical ventilation.

\paragraph{Lab redundancy:} 
The term ${r_{t}}^{\mtext{info}}$ denotes the feedback from taking one or more lab tests with novel information. We quantify this by using the mean squared difference between the last observed value and predictive means from the MOGP as a proxy for the information available:
\begin{equation}
{r_{t}}^{\mtext{info}} = \sum_{\ell=1}^{L} max\left(0, g(\cdot) - c_\ell\right) \cdot \mathbbm{1}_{\bm{a_{t}}[\ell] = 1} \,, \text{ where } g(\cdot) = \left| \frac{m_{t}^{(\ell)} - y_{t}^{(\ell)}}{\sigma_{t}^{(\ell)}} \right|,
\end{equation}
where $\sigma_{t}^{\ell}$ is the normalization coefficient for lab $\ell$, and the parameter $c_\ell$ determines the minimum prediction error necessary to trigger a reward; in our experiments, this is set to the median prediction error for labs ordered in the training data. The larger the deviation from current forecasts, the higher the potential information gain, and in turn the reward if the lab is taken.

\paragraph{Lab cost:} 
The last term in the reward function, ${r_{t}}^{\mtext{cost}}$ adds a penalty whenever any test is ordered to reflect the effective ``cost'' of taking the lab at time $t$.
\begin{equation}
{r_{t}}^{\mtext{cost }} = \sum_{\ell=1}^{L} \exp\left(-\frac{\Delta^{(\ell)}_t}{\Gamma_\ell}\right)  \cdot
\mathbbm{1}_{\bm{a_{t}}[\ell] = 1},
\end{equation}
where $\Gamma_\ell$ is a decay factor that controls the how fast the cost decays with the time $\Delta_t$ elapsed since the last measurement. In our experiments, we set $\Gamma_\ell = 6\,\, \forall \ell \in L$.

\subsection{Learning optimal policies}
\label{learningpi}
Once we extract sequences of states, actions, and rewards from the ICU data, we can generate a dataset of one-step transition tuples of the form $\mathcal{D} = \{\langle s_t^n, a_t^n, s_{t+1}^n \rangle, r_t^n\}$, $n = 1... |\mathcal{D}|$. These tuples can then be used to learn an estimate of the Q-function, $\hat{Q}: \mathcal{S} \times \mathcal{A} \rightarrow \mathbb{R}^d$ ---where $d=4$ is the dimensionality of the reward function---to map a given state-action pair to a vector of expected cumulative rewards. Each element in the Q-vector represents the estimated value of that state-action pair according to a different objective. We learn this Q-function using a variant of Fitted Q-iteration (FQI) with extremely randomized trees \citep{ernst2005tree, prasad2017reinforcement}. FQI is a batch off-policy reinforcement learning algorithm that is well-suited to clinical applications where we have limited data and challenging state dynamics. 
The algorithm adapted here to handle vector-valued rewards is based on Pareto-optimal Fitted-Q \cite{lizotte2016multi}. 

In order to scale from the two-stage decision problem originally tackled to the much longer admission sequences here ($\geq 24$ time steps), we define a stricter pruning of actions: at each iteration we eliminate any \emph{dominated} actions for a given state---those actions that are outperformed by alternatives for all elements of the Q-function---and retain only the set $\Pi(s) = \{a: \nexists a'\, (\forall \,d,\,\, \hat{Q}_d(s, a) < \hat{Q}_d(s, a'))\}$ for each $s$. Actions are further filtered for \emph{consistency}: we might consider feature consistency to be defined as rewards being linear in each feature space \cite{lizotte2016multi}. Here, we relax this idea to filter out only those actions from policies that cannot be expressed by our chosen nonlinear tree-based classifier.
The function will still yield a non-deterministic policy (NDP) as, in most cases, there will not be a strictly optimal action that achieves the highest $Q_d$ for all $d$. In the following section, we suggest one possible approach for reducing the NDP to give a single best action for any given state based on practical considerations for this setting.

\begin{algorithm}[t]
  \SetAlgoLined
  \textbf{Input:\\} {One-step transitions $\mathcal{F} = \{\langle s_t^n, a_t^n, s_{t+1}^n\rangle, r_{t+1}^n\}_{n=1:|\mathcal{F}|}$; \\ 
  $\qquad $ Regression parameters $\theta$}; action space $\mathcal{A}$; subset size $N$ \\
  \textbf{Initialize} ${Q^{(0)}}(s_t,a_t) = \bm{0} \in \mathbb{R}^d \quad \forall s_t \in \mathcal{F}, \, a_t \in \mathcal{A}$ \\
  \For{iteration $k = 1 \rightarrow K$}{
  {Sample $\textit{subset}_N \sim \mathcal{F}$} ; initialize
  {$S \leftarrow []$} \\
  \For{$i \in \textit{subset}_N$ }{
  Generate set $\Pi(s_i)$ using $Q^{(k-1)}$ \\
  Initialize classification parameters $\phi$ \\
  $\phi \leftarrow \textit{classify}(s_{i}, a_{i})$ \\
    \For{$\pi_i \in \Pi:$}{
    $a' \leftarrow \pi_i(s_{i+1}) \cap \textit{predict}(s_{i+1}, \phi)$\\
    {${Q^{(k)}}(s_i, a_i) \leftarrow r_{i+1} + \gamma 
    Q^{(k-1)}(s_{i+1}, a')$
    }}
    {$S \leftarrow \textit{append}(S, \langle(s_i, a_i), \, {Q^{(k)}}(s_i, a_i) \rangle)$} 
  }
  $\theta \leftarrow \textit{regress}(S)$
   } \KwResult{$\theta$}
   \caption{Multi-Objective Fitted Q-iteration with strict pruning (MO-FQI)}
   \label{fqi}
\end{algorithm}

\section{Results}
Following the extraction of our 6,060 admissions and resampling in hourly intervals using the forecasting MOGP, we partitioned the cohort into training and test sets of 3,636 and 2,424 admissions respectively. This gave approximately 500,000 one-step transition tuples of the form $\langle s_t, a_t, s_{t+1}, r_t\rangle$ in the training set, and over 350,000 in the test set.
We then ran batched FQI with these samples for $200$ iterations with discount factor $\gamma=0.9$. Each iteration took 100,000 transitions, sampled from the training set, with probability inversely proportional to the frequency of the action in the tuple. The vector-valued outputs of estimated Q-function were then used to obtain a non-deterministic policy for each lab considered (Section \ref{learningpi}). We chose to collapse this set to a practical deterministic policy as follows: 
\begin{equation}
\Pi(s)=
\begin{cases}
  1, & \,\, \hat{Q}_{d}(s, a=0) < \hat{Q}_{d}(s, a=1) + \varepsilon_d, \quad \forall \,d \\
  0, & \text{otherwise.}
\end{cases}
\end{equation} 
In particular, a lab should be taken \emph{only if} the action is \emph{optimal}, or estimated to outperform no other actions for all objectives in the Q-function. This strong condition for ordering a lab is motivated by the fact that the one of our primary objectives here is to minimize unnecessary ordering; the variable $\varepsilon_d$ allows us to relax this for certain objectives if desired. For example, if cost is a softer constraint in our case, setting $\varepsilon_{cost} > 0$ is an intuitive way to specify this preference in the policy. In our experiments, we tuned $\varepsilon_{cost}$ such that the total number of recommended orders of each lab approximates the number of actual orders in the training set.

\begin{figure}[t]
    \centering
            \vspace{-2mm}
    \includegraphics[width=\textwidth]{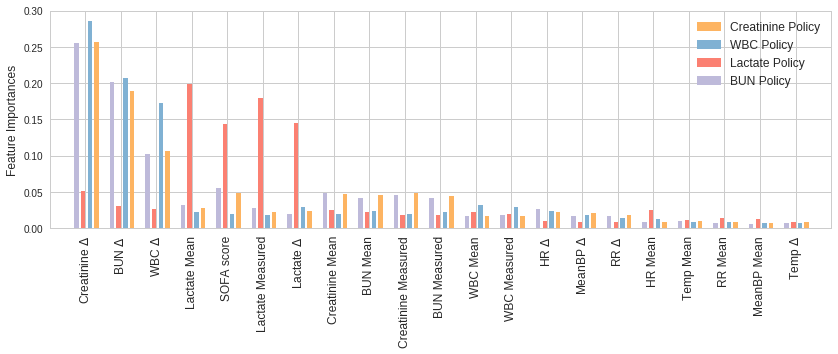}
    \vspace{-6mm}
    \caption{\textbf{Feature importances} over the 21-dimensional state space, for each of our four policies.}
        \vspace{-2mm}
\label{fig:featimportances}
\end{figure}

With a deterministic set of optimal actions, we could train our final policy function $\pi: \mathcal{S} \rightarrow{A}$; again, we used extremely randomized trees. The estimated feature importances of the policies learnt show that in the case of lactate the most important features are the mean and measured lactate, the time since last lactate measurement ($\Delta$) and the SOFA score (Figure \ref{fig:featimportances}). These relative importance scores are expected: a change in SOFA score may indicate the onset of sepsis, and in turn warrant a lactate test to confirm a source of infection, fitting typical clinical protocol. For the other three policies---WBC, creatinine, and BUN---again the time since last measurement of the respective lab tends be the prominent feature in the policy, along with the $\Delta$ terms for the other two labs. This emphasizes the overlap in information conveyed by these three tests: For example, abnormally high white blood cell count is a key criteria for sepsis, and severe sepsis often cascades into renal failure, which is typically diagnosed by elevated BUN and creatinine levels \cite{clarkson2010pocket}.

\begin{figure}[t]
    \centering
    \vspace{-5mm}
    \includegraphics[width=\textwidth]{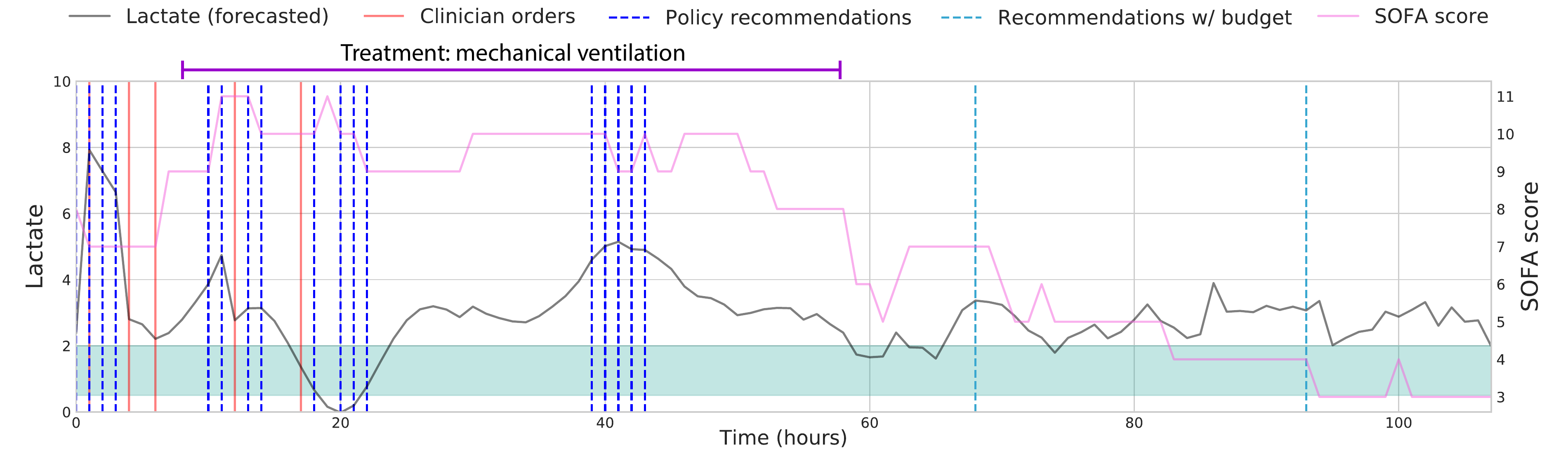}
    \vspace{-6mm}
    \caption{\textbf{Demonstration of one test trajectory of recommending lactate orders.} The shaded green region denotes the range of normal lactate values (0.5--2 mmol/L).}
\label{fig:test_traj_lactate}
\end{figure}

Once we have trained our policy functions, an additional component is added to our final recommendations: we introduce a \emph{budget} that suggests taking a lab at the end of every 24 hour period for which our policy recommends no orders. This allows us to handle regions of very sparse recommendations by the policy function, and reflects clinical protocols that require minimum daily monitoring of key labs.
In the policy for lactate orders in a typical patient admission, looking at the timing of the actual clinician orders, recommendations from our policy, and suggested orders from the budget framework, the actions are concentrated where lactate values are increasingly abnormal, or at sharp rises in SOFA score (Figure \ref{fig:test_traj_lactate}). 

\subsection{Off-Policy Evaluation}
We evaluated the quality of our final policy recommendations in a number of ways.
First, we implemented the per-step weighted importance sampling (PS-WIS) estimator to calculate the value of the policy $\pi_e$ to be evaluated:
\begin{equation*}
    \hat{V}_\text{\tiny{PS-WIS}}(\pi_e) = \sum^n_{i=1}\sum_{t=0}^{T-1}\gamma^t_{\text{\tiny{WIS}}}\left[\frac{\rho^{(i)}_t}{\sum^n_{i=1}\rho^{(i)}_t}\right]r^{(i)}_t, \quad \text{where } \rho_t = \prod^{t-1}_{j=0}\frac{\pi_e(s_j|a_j)}{\pi_b(s_j|a_j)},
\end{equation*}
given data collected from behaviour policy $\pi_b$ \cite{precup2000eligibility}. The behaviour policy was found by training a regressor on real state-action pairs observed in the dataset.
The discount factor was set to $\gamma_{\text{\tiny{WIS}}}=1.0$, so all time steps contribute equally to the value of a trajectory.

\begin{figure}[t]
    \centering
    \vspace{-5mm}
    \subfigure{\includegraphics[width=0.49\textwidth]{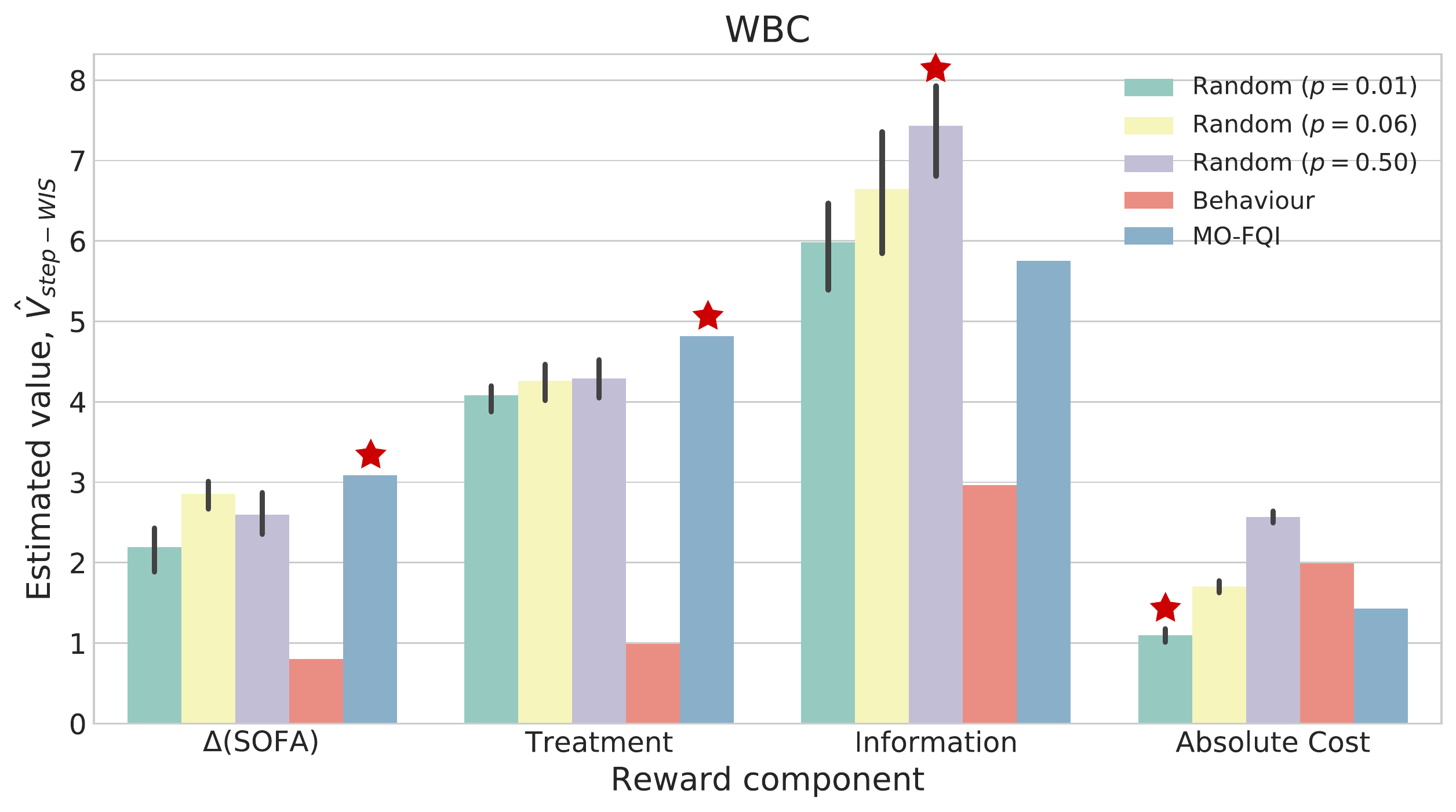}}
    \subfigure{\includegraphics[width=0.49\textwidth]{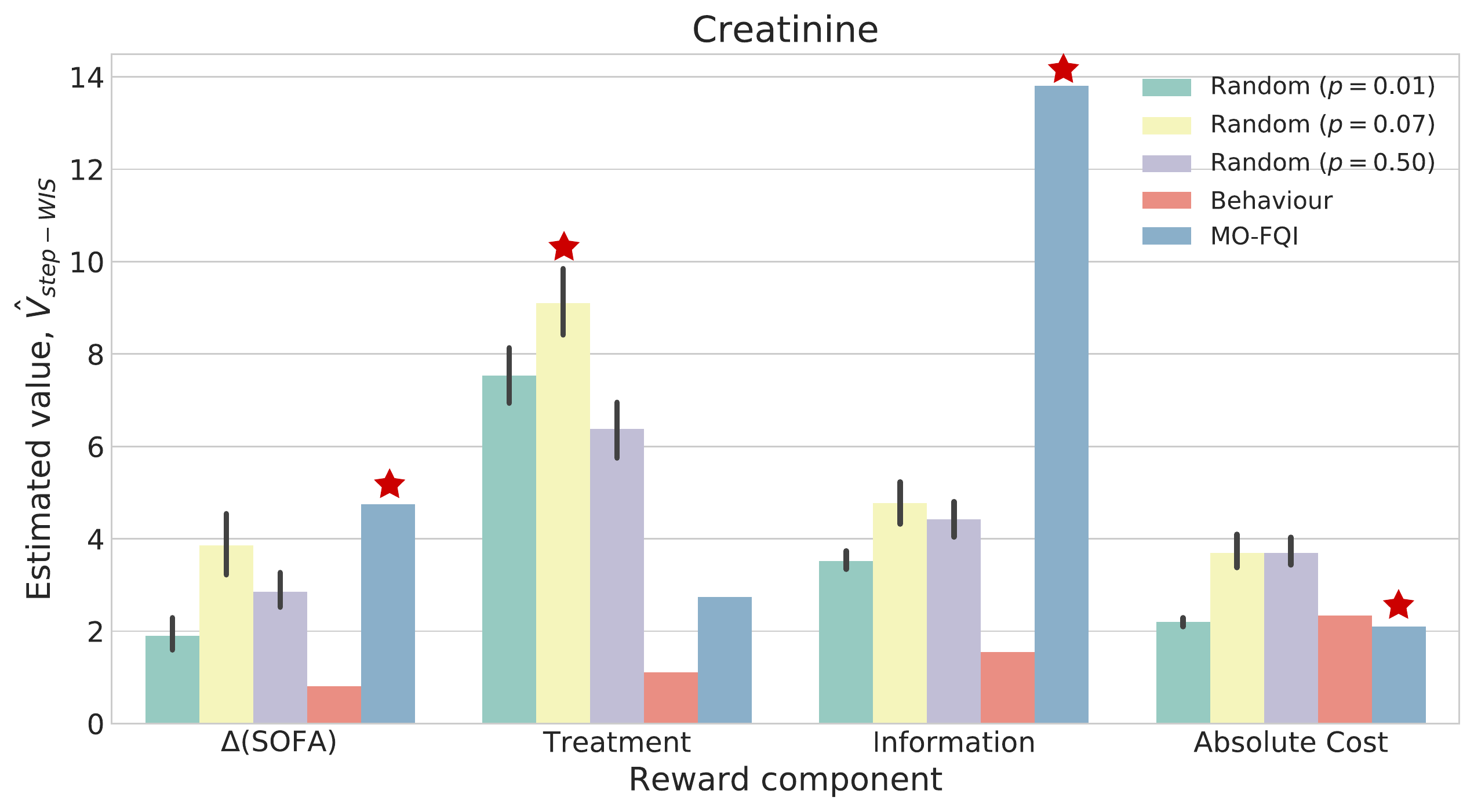}}
        \vspace{-3mm}
    \subfigure{\includegraphics[width=0.49\textwidth]{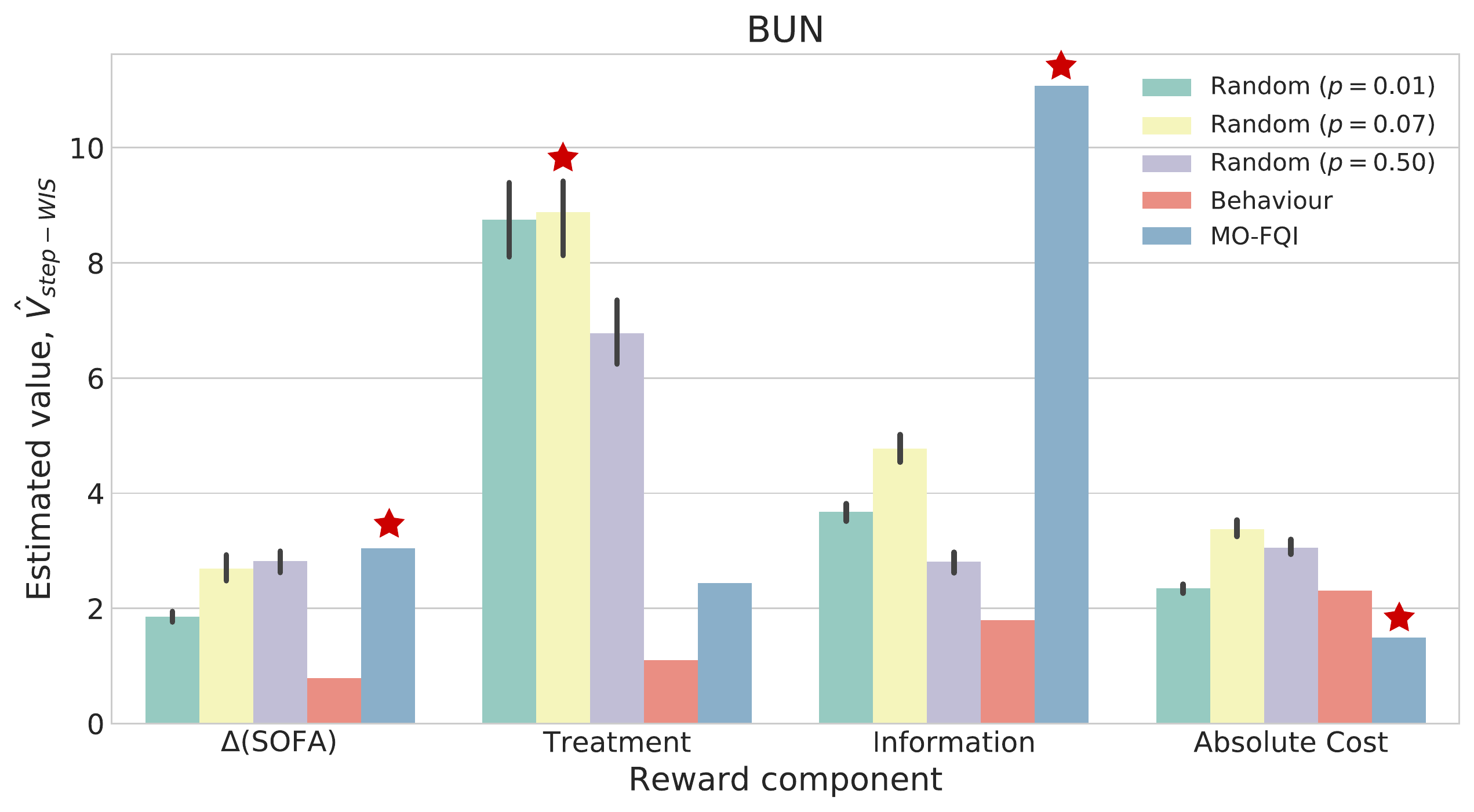}}
    \subfigure{\includegraphics[width=0.49\textwidth]{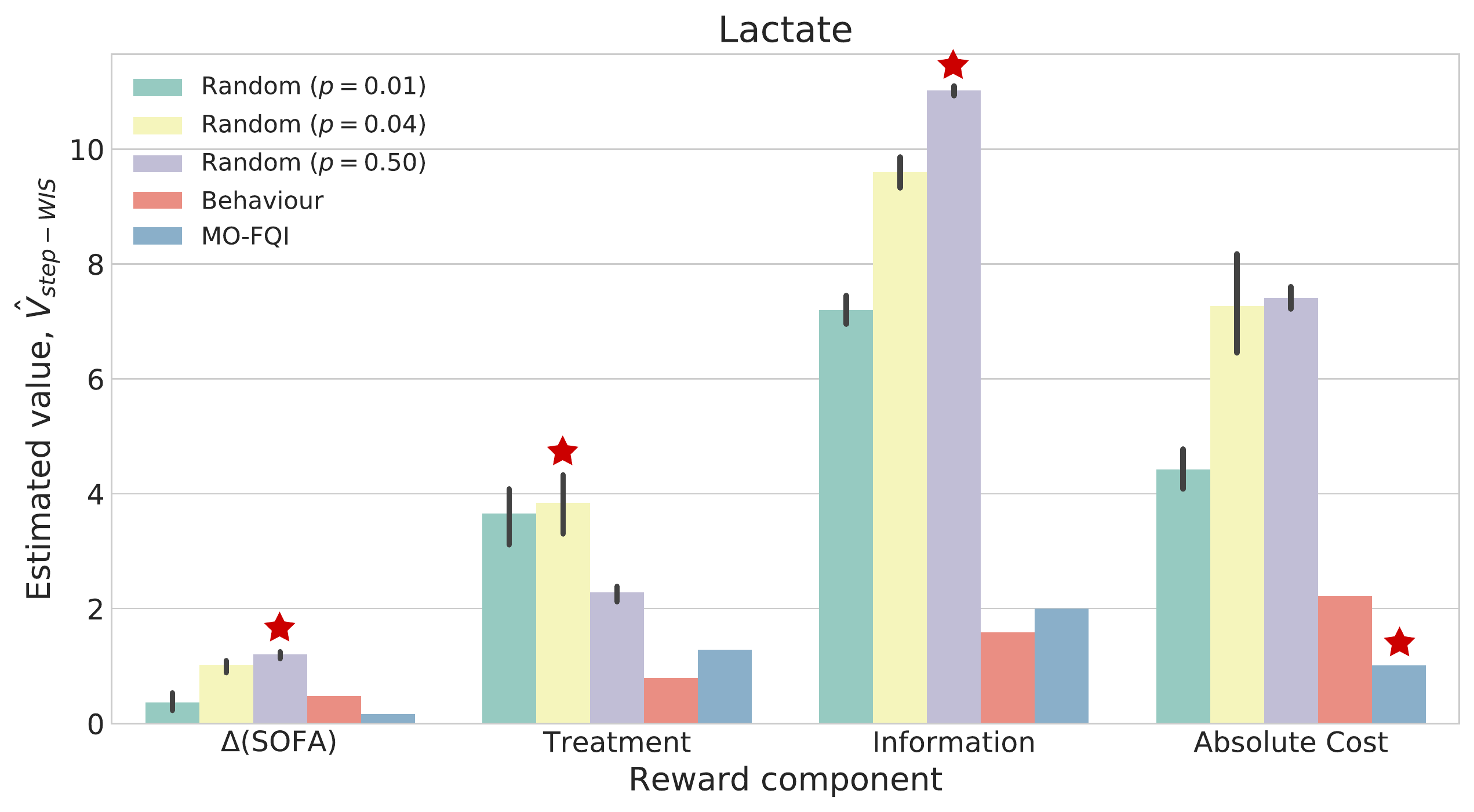}}
   \caption{\textbf{Evaluating $\bm{\hat{V}_d(\pi_e)}$} for each reward component $d$, across policies for four labs. For randomized policies, the error bars show the standard deviation across ten trials. The ({\color{Maroon}$\star$}) indicates the best performing policy for each reward component.}
    \label{fig:wdr}
            \vspace{-2mm}
\end{figure}

We then compared estimates for our policy (MO-FQI) against the behaviour policy and a set of randomized policies as baselines. These randomized policies were designed to generate random decisions to order a lab, with probabilities $p=\{0.01, p_{emp}, 0.5\}$, where $p_{emp}$ is the empirical probability of an order in the behaviour policy. For each $p$, we evaluated ten randomly generated policies and averaged performance over these. We observed that MO-FQI outperforms the behaviour policy across all reward components, for all four labs (Figure \ref{fig:wdr}). Our policy also consistently approximately matches or outperforms other policies in terms of cost---note that lower cost is better---even with the inclusion of the slack variable $\varepsilon_{cost}$ and the budget framework. Across the remaining objectives, MO-FQI outperforms the random policy in at least two of three components for all but lactate. This may be due in part to the relatively sparse orders for lactate resulting in higher variance value estimates.

In addition to evaluating using the per-step WIS estimator, we looked for more intuitive measures of how the final policy influences clinical practice. We computed three metrics here: (i) estimated reduction in total number of orders, (ii) mean information gain of orders taken, and (iii) time intervals between labs and subsequent treatment onsets.

\begin{figure}[t]
    \centering
    \vspace{-3mm}
     \subfigure{\includegraphics[width=0.48\textwidth]{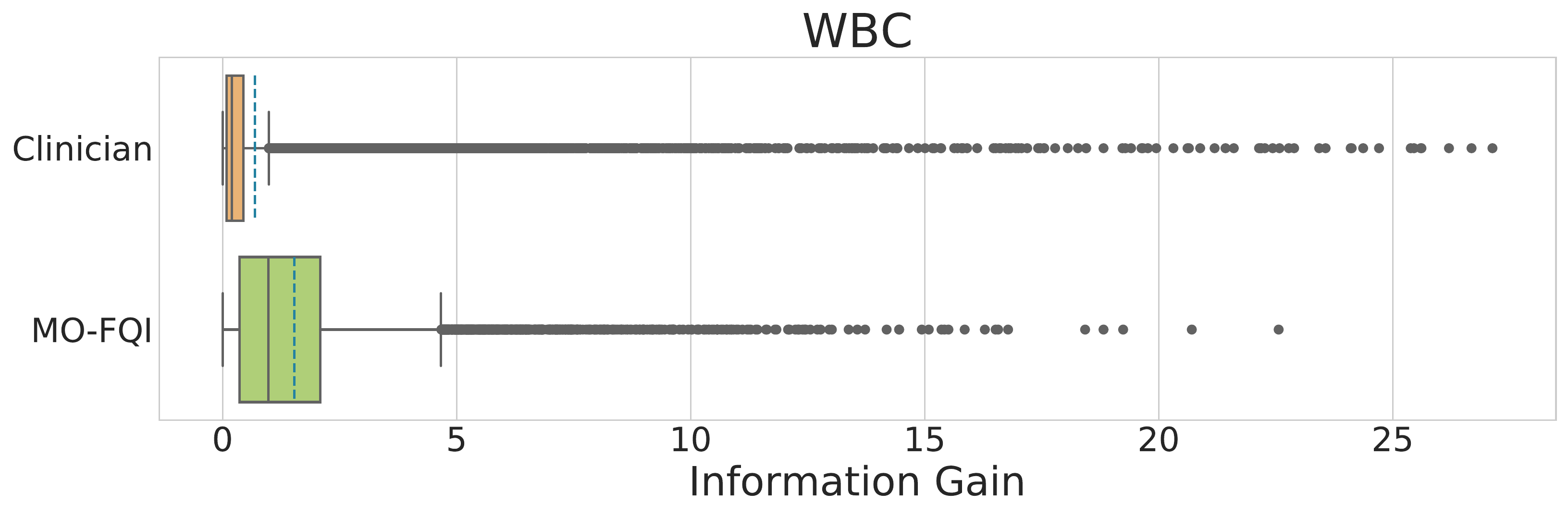}}
     \subfigure{\includegraphics[width=0.48\textwidth]{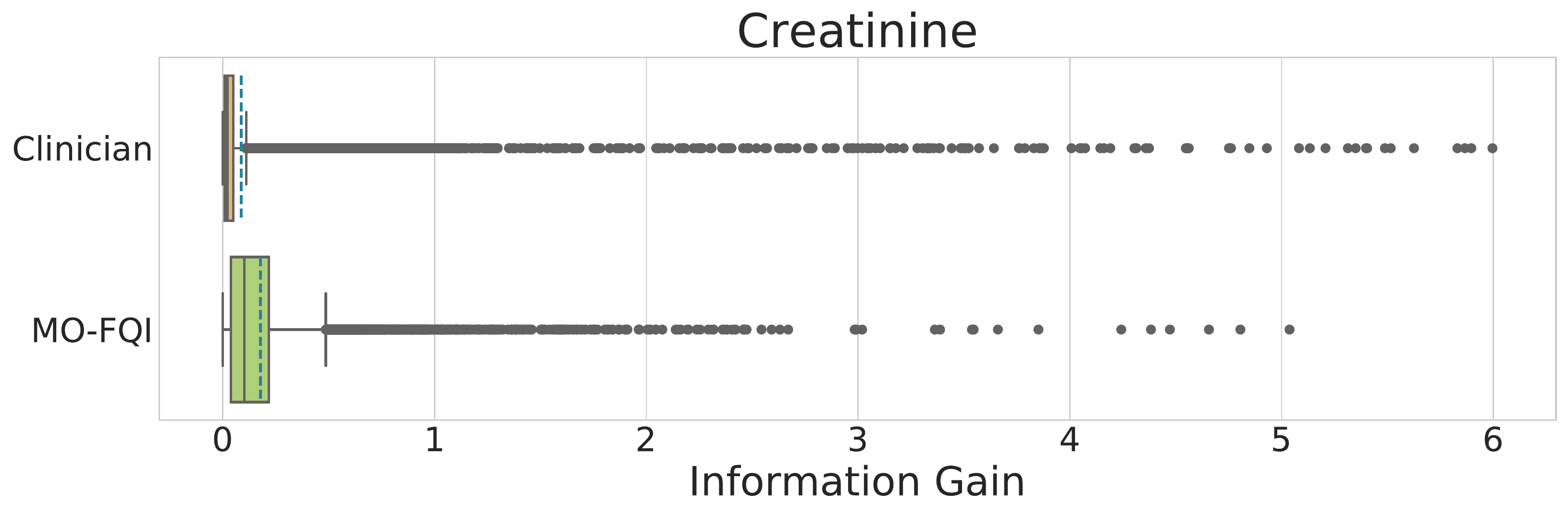}}\\
    \vspace{-3mm}
     \subfigure{\includegraphics[width=0.48\textwidth]{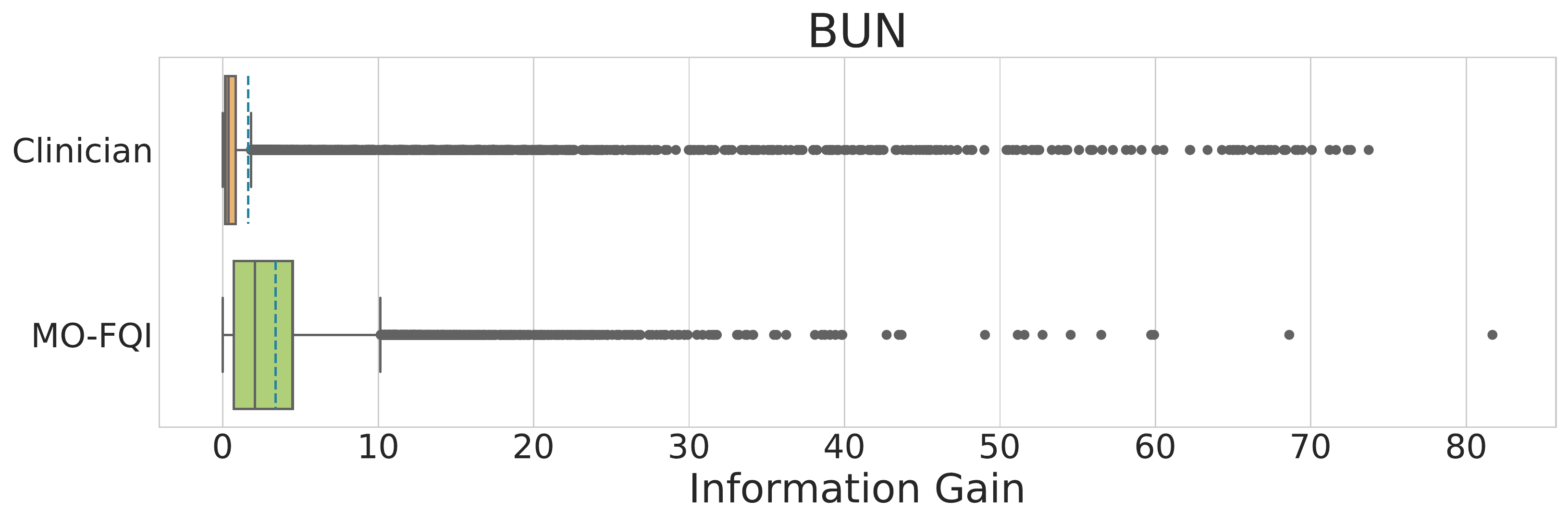}}
     \subfigure{\includegraphics[width=0.48\textwidth]{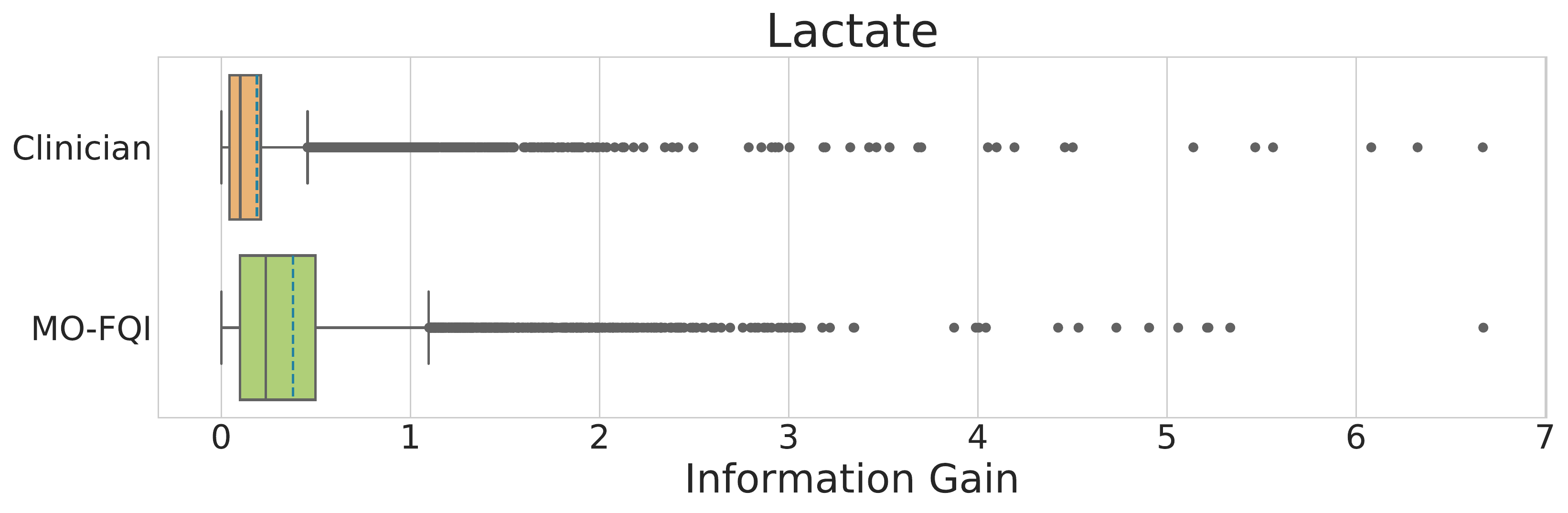}}
        \caption{\textbf{Evaluating Information Gain} of clinician actions against MO-FQI across all labs: the mean information in labs ordered by clinicians is consistently outperformed by MO-FQI: $0.69$ vs $1.53$ for WBC; $0.09$ vs $0.18$ for creatinine; $1.63$ vs $3.39$ for BUN; $0.19$ vs $0.38$ for lactate.
        } 
\label{fig:ig}
\end{figure}
\begin{figure}[t]
    \centering
    \vspace{-5mm}
    \subfigure{\includegraphics[width=0.48\textwidth]{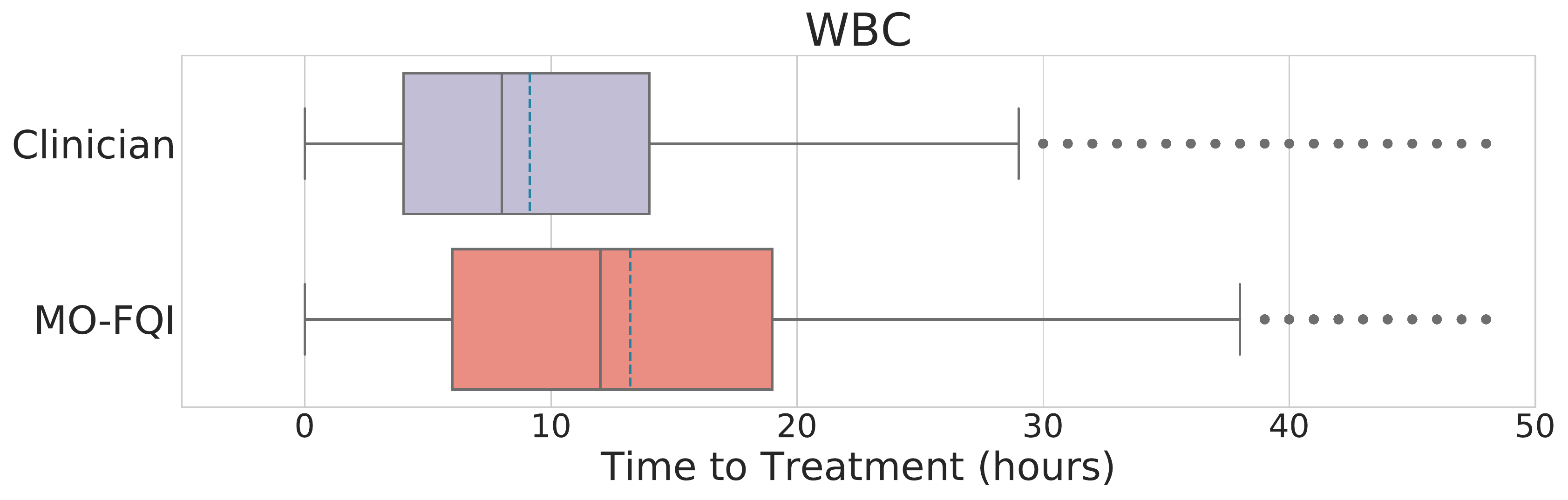}}
    \subfigure{\includegraphics[width=0.48\textwidth]{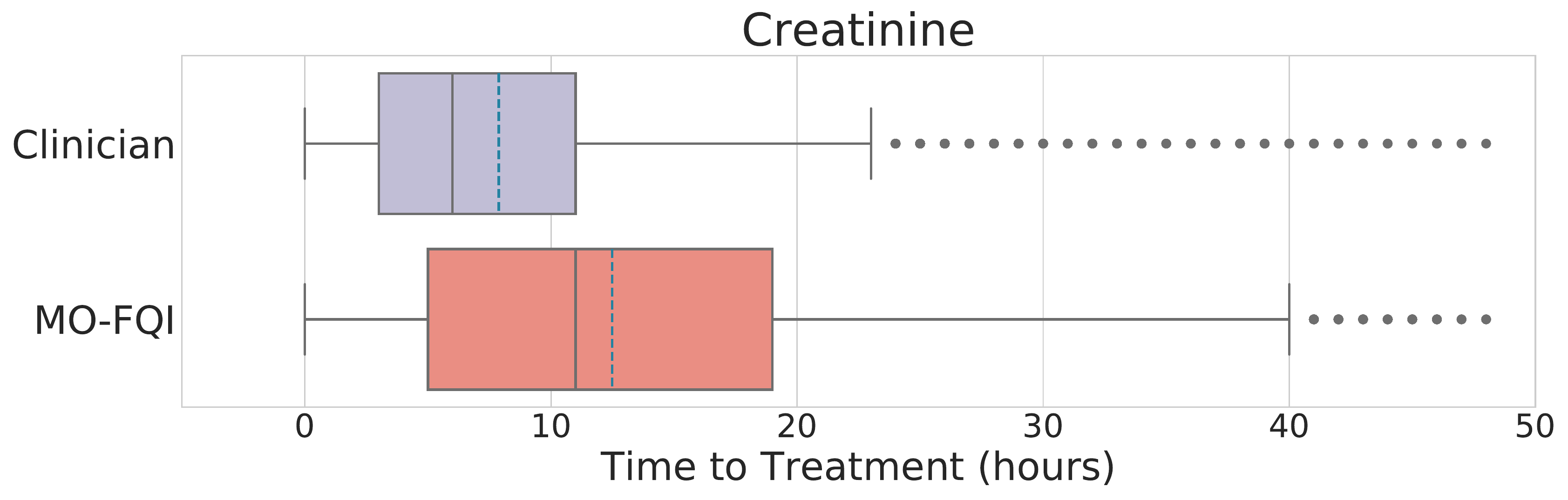}}\\
    \vspace{-3mm}
    \subfigure{\includegraphics[width=0.48\textwidth]{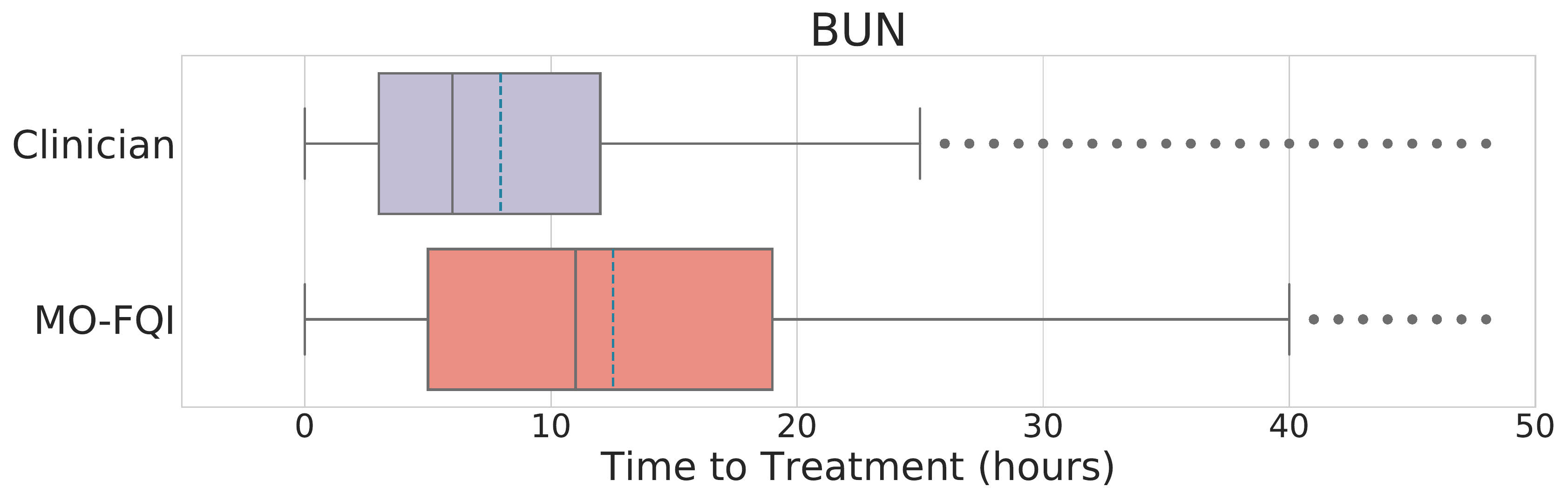}}
    \subfigure{\includegraphics[width=0.48\textwidth]{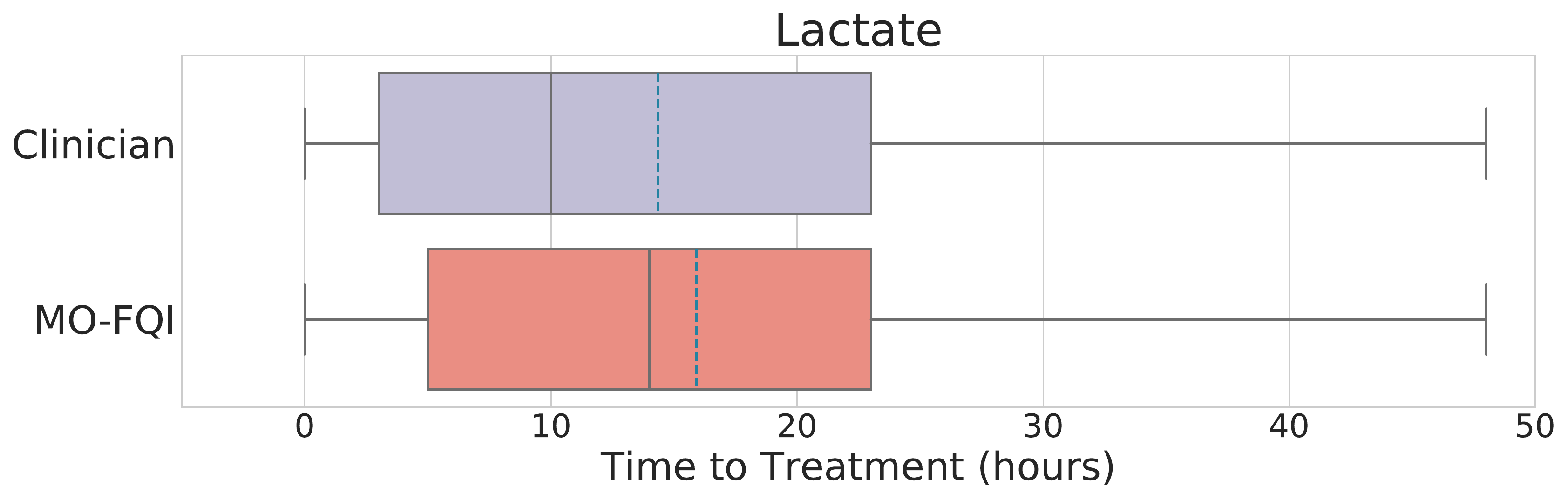}}
    \caption{\textbf{Evaluating Time to Treatment Onset} of lab orders by the clinician against MO-FQI across all labs: the mean time intervals are as follows (Clinician vs MO-FQI): $9.1$ vs $13.2$ for WBC; $7.9$ vs $12.5$ for creatinine; $8.0$ vs $12.5$ for BUN; $14.4$ vs $15.9$ for lactate.
    }
    \label{fig:tta}
\end{figure}

In evaluating the total number of recommended orders, we first filter a sequence of recommended orders to the just the first (onset) of recommendations if there are no clinician orders between them. We argue that this is a fair comparison as subsequent recommendations are made without counterfactual state estimation, i.e., without assuming that the first recommendation was followed the clinician. Empirically, we find that the total number of recommendations is considerably reduced. For instance, in the case of recommending WBC orders, our final policy reports 12,358 orders in the test set, achieving a reduction of 44\% from the number of true orders (22,172). In the case of lactate, for which clinicians' orders are the least frequent (14,558), we still achieved a reduction of 27\%.

We also compared the approximate information gain of the actions taken by the estimated policy, in comparison with the policy used in the collected data. To do this, we defined the information gain at a given time by looking at the difference between the \emph{approximated} true value of the target lab, which we impute using the MOGP model given all the observed values, and the forecasted value, computed using only the values observed before the current time.  The distribution of aggregate information gain for orders recommended by our policy and actual clinician's orders in the test set shows higher mean information gain with MO-FQI (Figure \ref{fig:ig}).

Lastly, we considered the time to onset of critical interventions, which we define to include initiation of vasopressors, antibiotics, mechanical ventilation or dialysis. We first obtained a sequence of treatment onset times for each test patient; for each of these time points, we traced back to the earliest observed or recommended order taking place within the past 48 hours, and computed the time between these: $\Delta_{t} = t_{\mtext{treatment}}-t_{\mtext{order}}$.  The distribution of time-to-treatment for labs taken by the clinician in the true trajectory against that for recommendations from our policy, for all four labs, shows that the recommended orders tend to happen earlier than the actual time of an order by the clinician---on average over an hour in advance for lactate, and more that four hours in advance for WBC, creatinine, and BUN (Figure \ref{fig:tta}).

\section{Conclusion}
In this work, we propose a reinforcement learning framework for decision support in the ICU that learns a compositional optimal treatment policy for the ordering of lab tests from sub-optimal histories. We do this by designing a multi-objective reward function that reflects clinical considerations when ordering labs, and adapting methods for multi-objective batch RL to learning extended sequences of Pareto-optimal actions. Our final policies are evaluated using importance-sampling based estimators for off-policy evaluation, metrics for improvements in cost, and reducing redundancy of orders. Our results suggest that there is considerable room for improvement on current ordering practices, and the framework introduced here can help recommend best practices and be used to evaluate deviations from these across care providers, driving us towards more efficient health care. Furthermore, the low risk of these types of interventions in patient health care reduces the barrier of testing and deploying clinician-in-the-loop machine learning-assisted patient care in ICU settings.

\bibliographystyle{ws-procs11x85}
\bibliography{reference}

\begin{thebibliography}{10}

\bibitem{badrick2013evidence}
T.~Badrick, Evidence-based laboratory medicine {\em The Clinical Biochemist
  Reviews} {\bf 34} (The Australian Association of Clinical Biochemists, 2013).

\bibitem{zhi2013landscape}
M.~Zhi, E.~L. Ding, J.~Theisen-Toupal, J.~Whelan and R.~Arnaout, The landscape
  of inappropriate laboratory testing: a 15-year meta-analysis {\em PloS one}
  {\bf 8} (Public Library of Science, 2013).

\bibitem{loftsgard2016clinicians}
T.~Loftsgard and R.~Kashyap, Clinicians role in reducing lab order frequency in
  icu settings {\em J Perioper Crit Intensive Care Nurs} {\bf 2}2016.

\bibitem{konger2016reduction}
R.~L. Konger, P.~Ndekwe, G.~Jones, R.~P. Schmidt, M.~Trey, E.~J. Baty,
  D.~Wilhite, I.~A. Munshi, B.~M. Sutter, M.~Rao {\em et~al.}, Reduction in
  unnecessary clinical laboratory testing through utilization management at a
  us government veterans affairs hospital {\em American journal of clinical
  pathology} {\bf 145} (Oxford University Press, 2016).

\bibitem{icumedical2015}
ICUMedical, Reducing the risk of iatrogenic anemia and catheter-related
  bloodstream infections using closed blood sampling (ICU Medical Inc., 2015).

\bibitem{lee2015using}
J.~Lee and D.~M. Maslove, Using information theory to identify redundancy in
  common laboratory tests in the intensive care unit {\em BMC medical
  informatics and decision making} {\bf 15} (BioMed Central, 2015).

\bibitem{cismondi2013reducing}
F.~Cismondi, L.~A. Celi, A.~S. Fialho, S.~M. Vieira, S.~R. Reti, J.~M. Sousa
  and S.~N. Finkelstein, Reducing unnecessary lab testing in the icu with
  artificial intelligence {\em International journal of medical informatics}
  {\bf 82} (Elsevier, 2013).

\bibitem{luosol2016}
Y.~Luo, P.~Szolovits, A.~S. Dighe and J.~M. Baron, Using machine learning to
  predict laboratory test results {\em American Journal of Clinical Pathology}
  {\bf 145}2016.

\bibitem{ghassemi2015multivariate}
M.~Ghassemi, M.~A.~F. Pimentel, T.~Naumann, T.~Brennan, D.~A. Clifton,
  P.~Szolovits and M.~Feng, A multivariate timeseries modeling approach to
  severity of illness assessment and forecasting in {ICU} with sparse,
  heterogeneous clinical data, in {\em Proceedings of the Twenty-Ninth AAAI
  Conference on Artificial Intelligence\/}, 2015.

\bibitem{Cheng2017arXiv}
L.-F. {Cheng}, G.~{Darnell}, C.~{Chivers}, M.~{Draugelis}, K.~{Li} and
  B.~{Engelhardt}, Sparse multi-output gaussian processes for medical time
  series prediction  (2017).

\bibitem{nemati2016optimal}
S.~Nemati, M.~M. Ghassemi and G.~D. Clifford, Optimal medication dosing from
  suboptimal clinical examples: A deep reinforcement learning approach, in {\em
  Engineering in Medicine and Biology Society (EMBC), 2016 IEEE 38th Annual
  International Conference of the\/}, 2016.

\bibitem{raghu2017continuous}
A.~Raghu, M.~Komorowski, L.~A. Celi, P.~Szolovits and M.~Ghassemi, Continuous
  state-space models for optimal sepsis treatment: a deep reinforcement
  learning approach, in {\em Proceedings of the Machine Learning for Health
  Care, {MLHC} 2017, Boston, Massachusetts, USA, 18-19 August 2017\/}, 2017.

\bibitem{prasad2017reinforcement}
N.~Prasad, L.~Cheng, C.~Chivers, M.~Draugelis and B.~Engelhardt, A
  reinforcement learning approach to weaning of mechanical ventilation in
  intensive care units, in {\em Uncertainty in Artificial Intelligence 2017\/},
  1 2017.

\bibitem{lizotte2016multi}
D.~J. Lizotte and E.~B. Laber, Multi-objective markov decision processes for
  data-driven decision support {\em Journal of Machine Learning Research} {\bf
  17}2016.

\bibitem{johnson2016mimic}
A.~E. Johnson, T.~J. Pollard, L.~Shen, L.-w.~H. Lehman, M.~Feng, M.~Ghassemi,
  B.~Moody, P.~Szolovits, L.~A. Celi and R.~G. Mark, Mimic-iii, a freely
  accessible critical care database {\em Scientific data} {\bf 3} (Nature
  Publishing Group, 2016).

\bibitem{singer2016third}
M.~Singer, C.~S. Deutschman, C.~W. Seymour, M.~Shankar-Hari, D.~Annane,
  M.~Bauer, R.~Bellomo, G.~R. Bernard, J.-D. Chiche, C.~M. Coopersmith {\em
  et~al.}, The third international consensus definitions for sepsis and septic
  shock (sepsis-3) {\em Jama} {\bf 315} (American Medical Association, 2016).

\bibitem{vincent2016qsofa}
J.-L. Vincent, G.~S. Martin and M.~M. Levy, qsofa does not replace sirs in the
  definition of sepsis {\em Critical Care} {\bf 20} (BioMed Central, 2016).

\bibitem{ernst2005tree}
D.~Ernst, P.~Geurts and L.~Wehenkel, Tree-based batch mode reinforcement
  learning {\em Journal of Machine Learning Research} {\bf 6}2005.

\bibitem{clarkson2010pocket}
M.~R. Clarkson, B.~M. Brenner and C.~Magee, {\em Pocket Companion to Brenner
  and Rector's The Kidney E-Book} (Elsevier Health Sciences, 2010).

\bibitem{precup2000eligibility}
D.~Precup, R.~S. Sutton and S.~P. Singh, Eligibility traces for off-policy
  policy evaluation, in {\em Proceedings of the Seventeenth International
  Conference on Machine Learning\/}, ICML '00 (Morgan Kaufmann Publishers Inc.,
  San Francisco, CA, USA, 2000).

\end{thebibliography}
\end{document}